\documentclass{article}

\usepackage{graphicx}
\usepackage[dblblindworkshop, final]{neurips_2025}
\workshoptitle{Mechanistic Interpretability}

\usepackage[utf8]{inputenc} 
\usepackage[T1]{fontenc}    
\usepackage{hyperref}       
\usepackage{url}            
\usepackage{booktabs}       
\usepackage{amsfonts}       
\usepackage{nicefrac}       
\usepackage{microtype}      
\usepackage{xcolor}         

\title{Emerging Human-like Strategies for Semantic Memory Foraging in Large Language Models}

\author{
  Eric Lacosse\textsuperscript{1,2}, Mariana Duarte\textsuperscript{1}, Peter M. Todd\textsuperscript{3}, Daniel C. McNamee\textsuperscript{1,2}\\
  \textsuperscript{1}Champalimaud Research, Lisbon, Portugal \\
  \textsuperscript{2}Centre for Restorative Neurotechnology, Lisbon, Portugal \\
  \textsuperscript{3}Indiana University, Bloomington, IN, USA \\
    \texttt{eric.lacosse@research.fchampalimaud.org}
}

\begin{document}

\maketitle

\begin{abstract}
Both humans and Large Language Models (LLMs) store a vast repository of semantic memories.
In humans, efficient and strategic access to this memory store is a critical foundation for a variety of cognitive functions.
Such access has long been a focus of psychology and the computational mechanisms behind it are now well characterized.
Much of this understanding has been gleaned from a widely-used  neuropsychological and cognitive science assessment called the Semantic Fluency Task (SFT), which requires the generation of as many semantically constrained concepts as possible.
Our goal is to apply mechanistic interpretability techniques to bring greater rigor to the study of semantic memory foraging in LLMs. To this end, we present preliminary results examining SFT as a case study.
A central focus is on convergent and divergent patterns of generative memory search, which in humans play complementary strategic roles in efficient memory foraging. We show that these same behavioral signatures, critical to human performance on the SFT, also emerge as identifiable patterns in LLMs across distinct layers. Potentially, this analysis provides new insights into how LLMs may be adapted into closer cognitive alignment with humans, or alternatively, guided toward productive cognitive \emph{disalignment} to enhance complementary strengths in human–AI interaction.
  
\end{abstract}

\section{Introduction}

The inherent complexity of large decoder-based transformers, particularly in their ability to mimic some human cognitive functions, has led to a surge in research focused on interpreting and analyzing their internal workings \cite{sharkey_open_2025, venhoff_understanding_2025}.
We suggest that cognitive mechanistic interpretability offers a remarkable opportunity to create and test scientific theories of the human mind, where AI systems can function as rigorous scientific artifacts to explain and predict human behavior \cite{frank_cognitive_2025}.
Towards that end, the cognitive behavior we focus on in this work is the process of active memory search \cite{crowder_principles_2014}.
Active memory search, although at times seemingly effortless, involves complex operations for humans \cite{todd_cognitive_2012}.
Our goal here is to understand how properties that explain memory operations in humans can similarly be identified as specific, explainable mechanisms used by LLMs appearing to emulate these cognitive operations in humans.
For example, we activate relevant concepts in our mental lexicon, quickly retrieve their associated information, and then produce that output to achieve our objective.
Although it is tempting to see these LLMs relying on the same cognitive mechanisms that explain human behavior, i.e., reflectionism, they are operating with an entirely different cognitive architecture and may not be ``thinking'' like us at all. Instead, they may be reproducing the patterns of our language under a ``strange and alien'' type of intelligence \cite{bratton_benjamin_after_nodate}, offering untapped possibilities for collaborative interaction \cite{schut_bridging_2023, sucholutsky_getting_2023}.
Disambiguating what cognitive mechanisms LLMs may be relying on is therefore needed.

\subsection{Semantic Fluency and Cognitive Search}
To understand active memory search in LLMs, we use the well-established Semantic Fluency Task (SFT) \cite{hills_optimal_2012}, a timed assessment measuring information retrieval from semantic memory to assess language production and executive functions, e.g., ``name all the animals you can in 3 minutes.'' 
During the SFT, humans ``cluster'' semantically or phonetically related words---\emph{convergent behavior}---and ``switch'' to new clusters when retrieval slows---\emph{divergent behavior} \cite{troyer_clustering_1997}.
This strategy is characterized by shorter pauses within clusters and longer ones between them \cite{hills_optimal_2012}.
Theoretically, this explore-exploit/converge-diverge behavior is modeled by the Marginal Value Theorem (MVT) of optimal foraging theory \cite{charnov_optimal_1976} and recent neural evidence for strategically timed switches in memory search are also further supported \cite{nour_trajectories_2023, lundin_neural_2023}.

\subsection{Prior Work and Our Contributions}

Previous studies on SFT have mainly focused on two areas: (1) using models to simulate and predict how humans generate words \cite{heineman_towards_2024}, and (2) using those models to identify patterns in human generated sequences \cite{hills_foraging_2015}.
In contrast, our study focuses specifically on how internal representations and mechanisms in LLMs can explain the cognitive dynamic of semantic foraging.
Towards this end, previous work has found that the distributions of attention weights within the LLM's layers and attention head may explain their semantic foraging behavior, specifically showing what LLM-based metrics can separate clustering and switching behavior given a human sequence \cite{zarries_components_nodate}. 
Our contributions focus on characterizing patterns of generative convergence versus divergence specifically within the layer dynamics of the LLMs' distributional readouts with a focus towards eventual steering applications. In this work:
\begin{itemize}
    \item  We analyzed a large dataset of semantic memory sequences from humans and language models, finding that LLM semantic search exhibits human-like patterns of convergent and divergent generation consistent with optimal memory search. A key distinction, however, is that LLMs exploit clusters more thoroughly, switching categories less frequently than their human counterparts.
    \item We show these mechanisms of semantic search behavior are identifiable in the token distribution patterns and the intermediate representational spaces across LLMs of various sizes. Specifically, we use logitlens and residual stream probing to trace where and how convergent and divergent semantic search behaviors emerge.
\end{itemize}

\section{Analysis and Results}

\subsection{Human versus LLM Semantic Fluency Behavior}
\label{sec:humanvsmachine}
\begin{figure*}
    \centering
    \includegraphics[width=\textwidth]{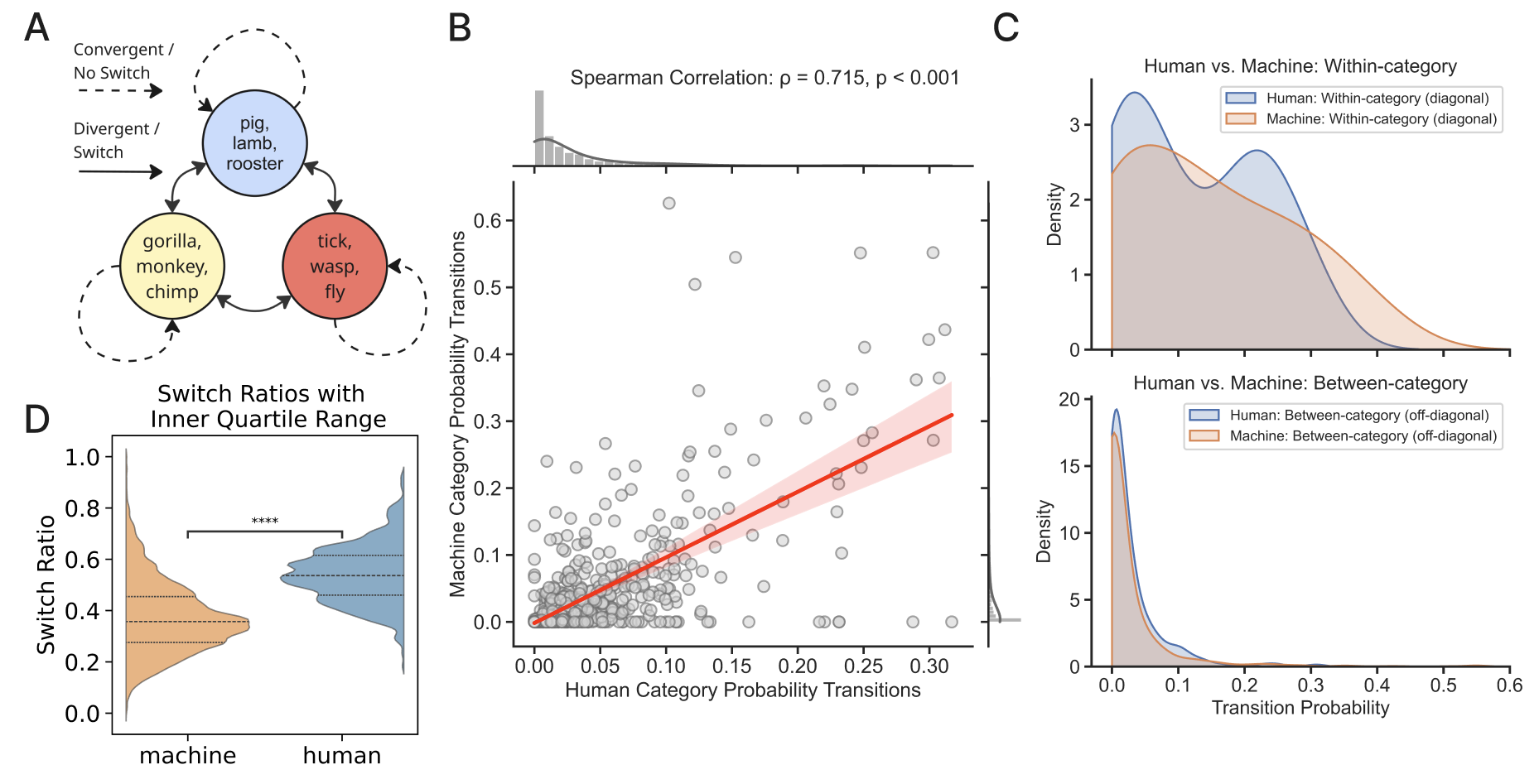}
    \caption{LLM and Human generated SFT sequences are compared. \textbf{A.} A state-transition diagram is drawn as conceptual illustration of three categories (non-human primates, insects, farm animals) that describe how a sequence is thought to be generated. Arrows between the nodes represent \emph{divergent} behavior whereas self-edges represent \emph{convergent}. Clustering refers to generating words within a specific category, while switching involves moving to a new category. \textbf{B.} The correlation between the average state-transition matrix representing transition probabilities between categories for human and LLM. \textbf{C.} LLM and human between-category and within-category transition probability distributions compared. \textbf{D.} Switch ratio distributions of human and LLM sequences showing that humans switch more often (mean switch ratio 0.55) than LLMs (mean switch ratio 0.4) indicating that LLMs are more effective at exhaustively sampling semantic clusters before switching. Mann-Whitney $p < 0.0001$.}
    \label{fig:1}
\end{figure*}

How do human and LLM generated sequences compare during semantic foraging?
To illustrate that LLM outputs largely resemble human sequence generation, Figure \ref{fig:1} shows similarities in human and LLM generated \emph{animal} sequences after both were instructed to perform an SFT task.

We analyzed 699 human-generated sequences of animal names, collected from three separate experiments where participants listed as many animals as they could in three minutes \cite{zemla_evidence_2023, hills_optimal_2012}.
We then generated an equal number of sequences (699) using the instruction-tuned Llama-3 suite of models (sizes 1B, 3B, 8B, and 70B) \cite{grattafiori_llama_2024}. 
To ensure variety, each LLM sequence began with the first word of a human sequence, using the following minimally sufficient user prompt that did not hold any explicit instruction to act like a human: 

\noindent\fbox{%
    \parbox{\dimexpr\textwidth-2\fboxsep-2\fboxrule\relax}{%
    \texttt{Without repeating yourself, continue your response as a list of comma separated animal names that come to mind: <Animal$_1$>,  [...GENERATION...]}%
    }%
}

All LLM generated sequences were continued until they were at least the same length as the human generated ones and then truncated to be the same number of responses afterwards: 35.
All responses, both human and LLM, were filtered according to strict criteria of needing to exist as a valid animal name defined by the category norms.
Any responses in generated sequences that were found to be invalid by the filter were grounds for discarding the entire sequence.
After filtering, a total of 681 sequences were analyzed for humans and 2285 sequences for LLMs (606, 502, 572, 605) for model sizes (1B, 3B, 8B, 70B), respectively.

In order to summarize generation patterns at the distributional level, a transition probability matrix was calculated for all human and machine sequences.
This models the state-transition diagram where nodes are (animal) categories and edges model their transition probabilities to either stay in the same category or switch to a different one (Figure \ref{fig:1}A).
Here we consider categories to be different \emph{animal} categories, e.g., farm animals, pets, etc.
Computing the Spearman correlation coefficient between averaged machine and human transition probability matrices reveals a strong correlation, $\rho = 0.701, p < 0.001$ (Figure \ref{fig:1}B).
Investigating whether humans and LLM tended to cluster their generation in the same categories, we saw a clear indication they both more likely transition within the same category as evidenced by the distribution of probabilities of the off-diagonal (between-categories) versus the diagonal (within-categories) matrix elements (Figure \ref{fig:1}C).
Switch ratios, the proportion of category switches for an individual's sequence, revealed that even though LLMs tend to match overall transition patterns to humans, LLMs tend to switch less often than humans (Figure \ref{fig:1}D).
These results demonstrate an overall \emph{macro-cognitive alignment}, i.e., population level alignment in cognitive behavioral patterns, between human and machine.

\subsection{Mapping Convergent and Divergent SFT Behavior}

\subsubsection{Investigating Output and Layer-wise Distributions}

\begin{figure*}
    \centering
    \includegraphics[width=\textwidth]{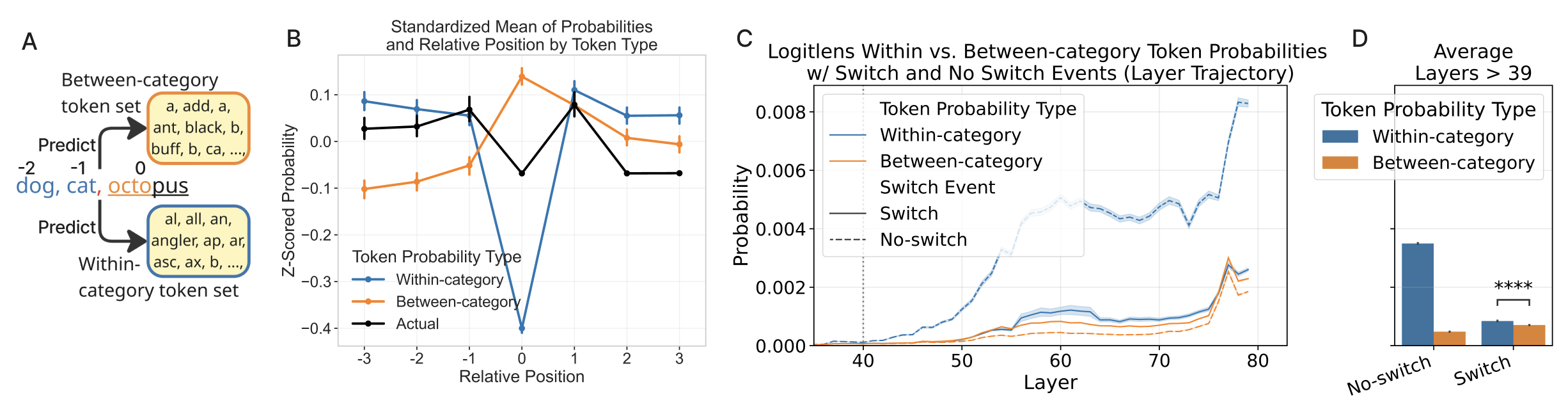}
    \caption{Explaining switching or convergent/divergent behavior in LLMs. \textbf{A.} Schematic illustrating the analysis of a switch event. For the preceding context, the model's vocabulary is partitioned into a Within-category token set (tokens that would continue the current `pet' cluster) and a Between-category token set (tokens that would initiate a switch to a new cluster) via animal category norms \cite{zemla_snafu_2020} \textbf{B.} Final output probabilities (z-scored) centered on the switch event (Relative Position 0). At the switch, the probability mass for Between-category tokens (orange) peaks, while the mass for Within-category tokens (blue) troughs. \textbf{C.} Logitlens analysis comparing probabilities during switch events (solid lines) vs. non-switch, convergent events (dashed lines). A clear separation emerges in the mid-to-late layers beyond layer 40. Critically, during a switch event, the model's tendency to amplify Within-category probabilities are greatly attenuated (compare solid blue vs. dashed blue line), providing a distinct, internal computational signature for the divergent ``switch'' mechanism. \textbf{D.} Average probability of Within-category vs. Between-category for switch vs. non-switch events from layer $>$ 39 where token probabilities in the different categories examined begin to differ from 0, as shown in panel C.}
    \label{fig:2}
\end{figure*}

We investigated whether human behavior for this task can be evaluated by an LLM to detect their switch behavior from token probabilities alone.
To do so, we examine the model's output distributions as well as activations of the intermediate layers that led up to its final output via the logitlens method \cite{nostalgebraist_interpreting_2020}.
First, in order to better understand the dynamics of these computations, for each animal in the generated sequence, we distinguished the set of tokens that are \emph{between}-category and \emph{within}-category sets (Figure \ref{fig:2}A).
These sets are defined by the category norms, used to compute the transition-probability matrix \ref{sec:humanvsmachine} (See details \ref{supp:norms}).
Figure \ref{fig:2}A shows how in the sequence ``dog, cat, octopus,'' dog and cat occupy relative positions $-2, -1$ (before) a switch event because the next animal octopus (relative position $0$) represents a switch to a different category, i.e., dog and cat belong in the category ``pets,'' whereas octopus belongs in the category ``sea creature.''
Probability estimates of the next-token are conditioned on sequence tokens making up cat, dog and the comma immediately after to determine the next-token probability estimates.
The output probability mass falls over both between-category sets and within-category sets.
The token vocabulary is parsed into the two set types and the probabilities are averaged within them, respectively.

Figure \ref{fig:2}B reveals that the model examined (Llama-3.3-70B) measures switch behavior for the actual human sequences to be surprising, showing lower probability at switch events in the sequence for within-category and the actual (belonging in the sequence).
Comparing mean z-score probability values between relative positions -1 and 0 for each token distribution type using permutation tests (10,000 resamples) yields significant effects (within-category, $d = -0.158$; between-category, $d = 0.144$; actual sequence, $d = -0.184$; all $p<0.001$).

Properties found within intermediate layers are thought to provide insight into how computations may facilitate certain downstream tasks \cite{hosseini_large_2023}.
This internal step-by-step processing of transformers may even mirror the cognitive processing of humans \cite{hu_signatures_2025} and may encode richer representations \cite{skean_layer_2025}.
We therefore investigate within/between token distributional dynamics within the model's intermediate layers, applying logitlens to the 70B model \cite{nostalgebraist_interpreting_2020}.
Appearing from middle layers, the model begins to distinguish within/between token-set distributions and switch events differently (Figure \ref{fig:2}C).
Within-category probability estimates increase the most dramatically, likely reflecting the ``stickiness'' to generate exemplars from within the same categories. 
Within-category probabilities are significantly attenuated during category switching, congruent with the observations found in Figure \ref{fig:2}B.
The appearance of between-category probability estimates being \emph{above} estimates from non-switch behavior appears relatable to the relative increase of between-category probability estimates shown in Figure \ref{fig:2}B.
Figure \ref{fig:2}D summarizes activations for the late layers (> 39), where the model shows a clear distinction between switch vs. non-switch events.
During non-switch events (False), the model is strongly convergent, with high (averaged) Within-category probability (0.0035) and low Between-category probability (0.0005). 
Critically, during a switch event (Figure \ref{fig:2}D, True), this convergent bias is neutralized: the Within-category probability is suppressed to 0.0008, becoming very close to the Between-category probability (0.0007), although still statistically different (Mann-Whitney, two-sided, $p<0.001$).

\subsubsection{Investigating Residual-Stream Representation}

\begin{figure*}
    \centering
    \includegraphics[width=\textwidth]{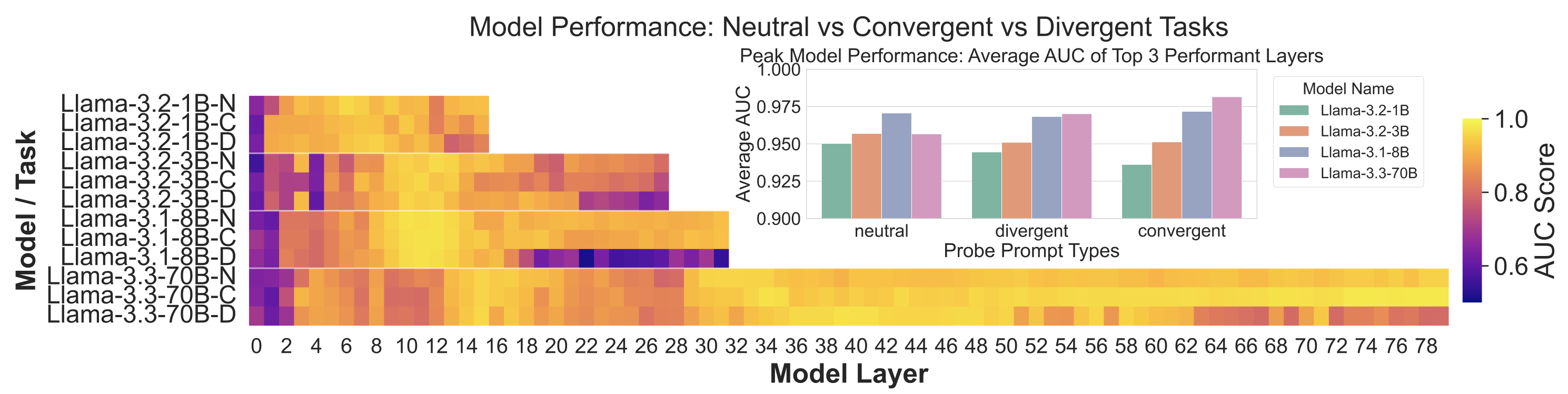}
    \caption{Decodability of Semantic Foraging Behavior from Internal Representations. The heatmap displays the layer-wise classification performance (AUROC) of linear probes trained to distinguish switch (divergent) vs. non-switch (convergent) events across four Llama models of increasing scale (1B, 3B, 8B, 70B) and a dataset of three different prompt types: N-neutral, C-convergent, and D-divergent. A summary of classifier performance for the averaged top-3 layers of the representation reading of the three different datasets demonstrates that increasing model size tends to improve performance.}
    \label{fig:3}
\end{figure*}

Can we identify switching behavior within the model representations themselves, probing the intermediate residual stream of the LLM \cite{belinkov_probing_2022}?
To do so, we trained a simple logistic regression classifier on the intermediate activations, which were first reduced using PCA.
All reported classifier performance is over an 80/20 cross-validation split.
First, we investigate whether human generated sequences are distinguishable from within these intermediate activations alone where switching behavior was labeled by category norms reported earlier.
This classifier was a very weak predictor ($\textsc{AUROC}=0.57$, Llama-3.3-70B) of switch events (Supplementary Figure S1), performing worse than a classifier from output distributions, showed earlier in Figure \ref{fig:2}C yielding an $\textsc{AUROC} = 0.751$ (Supplementary Figure S2) shown for different model sizes.

To explore this further, we exaggerated the semantic distances of switch or non-switch exemplars, amplifying either convergent behavior (staying within a category) or divergent behavior (switching to a new category).
This was done by creating contrastive pair datasets, bootstrapped from the human sequences evaluated earlier. 
Details of this procedure are available in supplements \ref{supp:contrastive}.
When classifiers were trained on representations from this new contrastive pair data, they performed exceptionally well at distinguishing between switch and non-switch events on residual-stream representations ($\textsc{AUROC}=0.96$, Llama-3.3-70B, neutral; $\textsc{AUROC}=0.98$, Llama-3.3-70B, convergence; $\textsc{AUROC}=0.97$, Llama-3.3-70B, divergence) (Figure \ref{fig:3}A).
A layer-wise analysis showed that the intermediate layers were the most effective for this classification (Figure \ref{fig:2}B). 
Interestingly, only in the divergent prompt instructed case, performance peaked in the middle layers and then dropped in the later layers, suggesting that when the model is instructed to produce divergent responses, representations of exemplars are moved closer together in later layers.

\section{Conclusion and Future Work}
Our work provides evidence that the strategic cognitive mechanisms of semantic foraging are an identifiable and potentially steerable property of LLMs \cite{turner_steering_2024}.
Future work will relate human processing details, e.g., time to generate responses, to the patterns we reported as we believe they may serve as features used to predict and explain properties of human cognition, allowing us to map internal model states to cognitive representations and computations.
Additionally, we seek to investigate whether a generalizable mechanism can be identified and rigorously tested.
These findings open new avenues for ``cognitive (dis)alignment'' in mechanistic interpretability research, where we may either strategically align models with patterns of human cognition or deliberately misalign them to inform and develop novel, more creative AI systems that may productively depart from our own cognitive behavior.
We hope to demonstrate that in applied collaborative contexts where approaches like representation engineering may offer an ability to steer model behavior in augmentative ways that extend beyond our own cognitive endowments \cite{zou_representation_2023}.

\bibliographystyle{plain} 
\bibliography{references}    

\newpage 
\appendix

\setcounter{figure}{0}
\setcounter{table}{0}
\setcounter{equation}{0}
\renewcommand{\thefigure}{S\arabic{figure}}
\renewcommand{\thetable}{S\arabic{table}}
\renewcommand{\theequation}{S\arabic{equation}}
\hrule
\section{Technical Appendices and Supplementary Material}

\subsection{Category Norms and Defining Switches}
\label{supp:norms}
In the context of cognitive psychology and neuropsychology, category norms are collections of data that represent the typical responses given by a group of people for a specific category.
Essentially, norms provide a baseline or a standard of comparison.
In this study, we evaluate the categories people give when they think of animals.
Using established category norms for animals \cite{zemla_snafu_2020}, we parsed all sequences to identify switch events as two sequential animals with no shared categories, i.e., divergences, and non-switch events, i.e., convergences as two sequential animals with at least one shared category.

\subsection{Contrastive Generation}
\label{supp:contrastive}
Three contrastive datasets were constructed to illustrate that we could effectively isolate a core mechanism for SFT.
We generated three datasets using distinct prompt instructions. 
In each case, \texttt{<SEQUENCE>} represents the placeholder for the preceding, 
comma-separated list of animals.
\subsection*{1. Neutral Prompt (Baseline)}
This prompt provided a general instruction without specific cognitive constraints.
\begin{quote}
    \texttt{Without repeating yourself, provide the next animal in the comma separated 
    list that comes immediately to mind: <SEQUENCE>,}
\end{quote}

\subsection*{2. Convergent Prompt (Clustering)}
This prompt explicitly instructed the model to stay within the current semantic category.
\begin{quote}
    \texttt{Without repeating yourself, provide the next animal in the comma separated 
    list that comes immediately to mind that sticks/stays/clusters/converges 
    to the same kind/type/category of last animal in the list: <SEQUENCE>,}
\end{quote}

\subsection*{3. Divergent Prompt (Switching)}
This prompt explicitly instructed the model to switch to a new, drastically 
different semantic category.
\begin{quote}
    \texttt{Without repeating yourself, provide the next animal in the comma separated 
    list that comes immediately to mind that diverges/moves/switches/changes 
    drastically away from the kind/type/category of last animal in the list: <SEQUENCE>,}
\end{quote}

We selected two sub-sequences, randomly sub-sampled from each human generated sequence, to include one non-switch (sub) sequence, and one switch (sub) sequence.
We replaced the next animal in that actual (sub) sequence to either be a maximally positive convergent/divergent example or a maximally negative convergent/divergent example by relying on an external embedding model\cite{pennington_glove_2014} and cosine similarity measure.
This allowed making a determination of what animal (belonging in the SNAFU defined norms \cite{zemla_snafu_2020}) would be either maximally/minimally convergent or divergent based on that cosine similarity measure.
These new question-and-answer pairs that prompt for either a neural, convergent, or a divergent response based on the sequence so far were used to readout the convergent or divergent behavior we report in the classifiers Figure \ref{fig:2} that significantly improved the ability to readout switch behavior.

\begin{figure*}
    \centering
    \includegraphics[width=15cm]{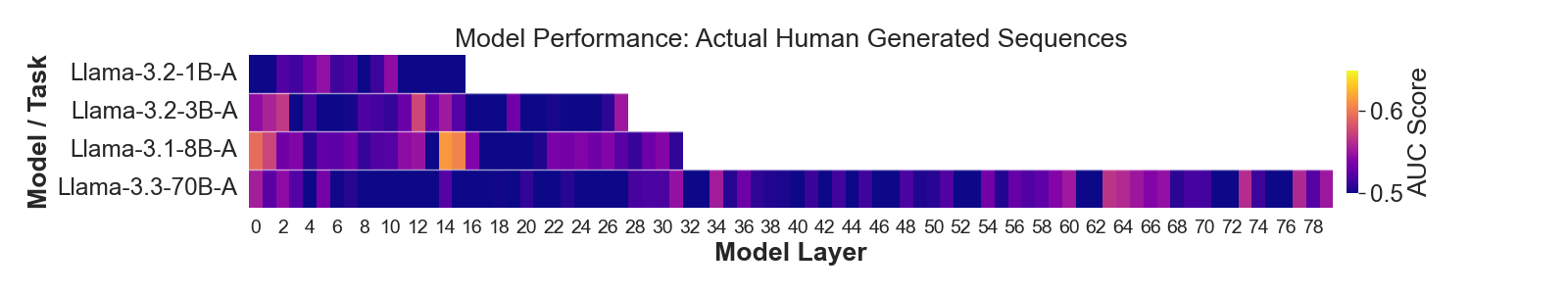}
    \caption{Investigating whether linear probing allows accurate readouts of human generated sequences from the human dataset, method akin to Figure \ref{fig:3}. The highest accuracies were achieved in Llama-3.1-8-A within layer 14, $\textsc{AUROC}=0.60$}.
    \label{fig:4}
\end{figure*}

\begin{figure*}
    \centering
    \includegraphics[width=10cm]{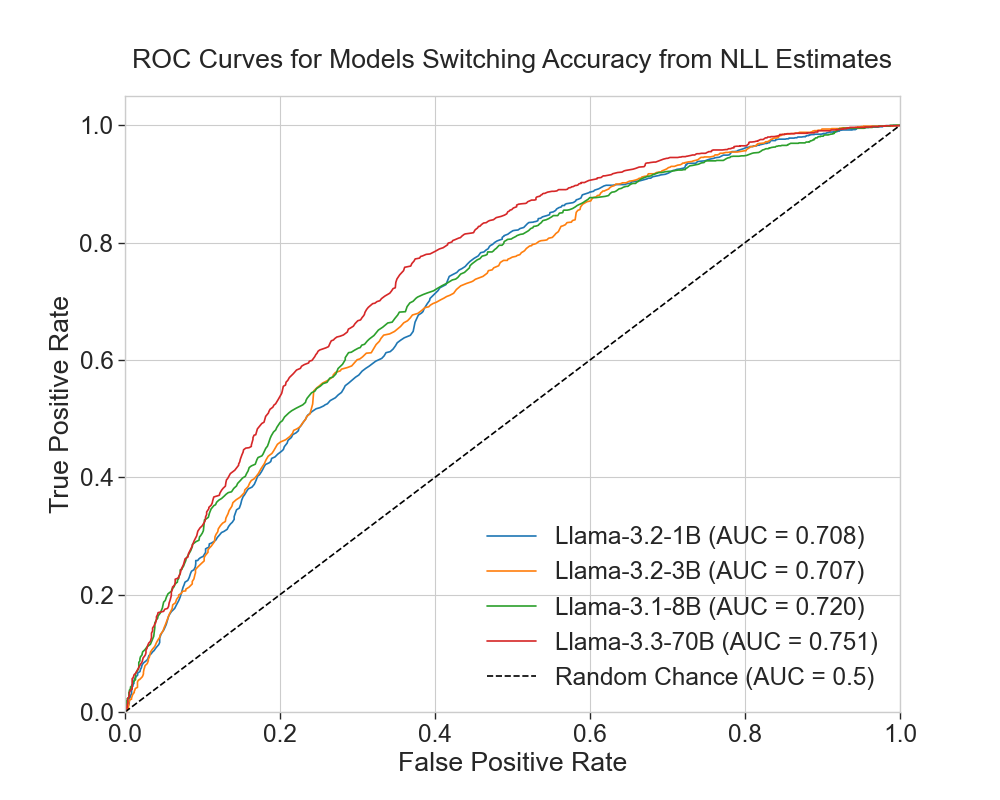}
    \caption{Classification of switching behavior from NLL output estimates for all model sizes for the actual token sequences of the human examined in Figure \ref{fig:2}. Larger models demonstrate higher accuracy.}
    \label{fig:5}
\end{figure*}

\end{document}